\documentclass[11pt,a4paper,twocolumn]{article}
\usepackage[T1]{fontenc}
\providecommand{\LyX}{L\kern-.1667em\lower.25em\hbox{Y}\kern-.125emX\@}

\newenvironment{lyxcode}
  {\begin{list}{}{
    \setlength{\rightmargin}{\leftmargin}
    \raggedright
    \setlength{\itemsep}{0pt}
    \setlength{\parsep}{0pt}
    \ttfamily}%
   \item[]}
  {\end{list}}

\begin{document}

\title{The role of robust semantic analysis in spoken language dialogue systems}

\author{Afzal Ballim and Vincenzo Pallotta\\
 MEDIA group: Laboratoire d'Informatique Théorique (LITH)\\
 École Polytechnique Fédérale de Lausanne (EPFL)\\
 IN-F Ecublens 1015 Lausanne (Switzerland)\\
 Phone:+41-21-693 52 97~~~ Fax:+41-21-693 52 78\\
 \texttt{\textbf{\small \{ballim,pallotta\}@di.epfl.ch}}\small }

\date{~}

\maketitle

\pagestyle{empty}
\thispagestyle{empty}

\section{Introduction}

The general applicative framework of the ISIS project\footnote{%
ISIS project started on April 1998 and finished on April 1999. It was funded
and overseen by SwissCom; the partners were EPFL (LIA and LITH), ISSCO and IDIAP.
More information can be found at \texttt{\footnotesize lithwww.epfl.ch/\~{}pallotta/rapportfinal.ps.gz}{\footnotesize \par}
} \cite{j-c.99:_isis_projec} was to design a NLP interface for automated telephone-based
phone-book inquiry. The objective of the project was to define an architecture
to improve speech recognition results by integrating higher level linguistic
knowledge. The availability of a huge collection of annotated telephone calls
for querying the Swiss phone-book database (i.e the Swiss French PolyPhone corpus)
allowed us to propose and evaluate a very first functional prototype of software
architecture for vocal access to database through phone and to test our recent
findings in semantic robust analysis obtained in the context of the Swiss National
Fund research project ROTA (Robust Text Analysis).

One of the main issues which has been taken into consideration is about robustness.
Robustness in dialogue is crucial when the artificial system takes part in the
interaction since inability or low performance in processing utterances will
cause unacceptable degradation of the overall system. As pointed out in \cite{Allen96:_robus_system_natur_spoken_dialog}
it is often better to have a dialogue system that tries to guess a specific
interpretation in case of ambiguity rather than ask the user for a clarification.
If this first commitment results later have to been a mistake, a robust behavior
will be able to interpret subsequent corrections as repair procedures to be
issued in order to get the intended interpretation.

\subsection{ISIS architecture}

Dialogue processing requires large amount of domain knowledge as well as linguistic
knowledge in order to ensure acceptable coverage and understanding. Cooperation
between processing modules and the integration of various knowledge resources
require the design of a suitable software architecture. In the ISIS project
the processing of the corpus data is performed at various linguistic levels
by modules organized into a pipeline. Each module assumes as input the output
of the preceding module. The main goal of this architecture is to understand
how far it is possible go without using any kind of feedback among different
linguistic modules. In this paper we will detail the functionality of the semantic
module.

\section{Robust semantic analysis}

In theory, a complete dialogue management system requires total semantic understanding
of the input. However, as we all know this is not possible with current systems,
and may never be possible. Even restricting ourselves to a limited domain it
is still very difficult to get any useful semantic representation from free
dialogue. 

A different approach considers that a dialogue management can be achieved by
a light parsing of the input. This method needs neither a full semantic understanding
of the language nor a deep investigation in the meaning and senses of the words.
It is merely based on the knowledge of certain \emph{cue-phrases} able to describe
a shallow semantic structure of the text. These cue-phrases or terms should
be relevant enough to give a coherent semantic separation of the different parts.
Nonetheless, the set of terms must not be rigid to avoid boolean results, but
there must be a set of semantically similar terms with a degree of confidence
for each. This would generate hypothetic semantic descriptions. In fact, these
terms which correspond to semantic fields, are able to isolate texts parts.
In case of a failure in obtaining a full and precise semantic description, a
minimum description would indeed be derived. Therefore in all cases only relevant
parts would undergo the understanding process. A similar approach has been proposed
by Grefenstette in \cite{g.96:_light_parsin_finit_state_filter} where the main
applications are slanted to the extraction of syntactic information (e.g. grouping
adjacent syntactically-related units and extracting non-adjacent n-ary grammatical
relations). Finite-states parsing technology has been adopted as a solution
in order to achieve robustness and efficiency at the implementation level. 

Although robustness can be considered as being applied at either a syntactic
or semantic level, we believe it is generally at the semantic level that it
is most effective. This robust analysis needs a model of the domain in which
the system operates, and a way of linking this model to the lexicon used by
the other components. It specifies semantic constraints that apply in the world
and which allow us, for instance, to rule out incoherent requests. The degree
of detail required of the domain model used by the robust analyzer depends upon
the ultimate task that must be performed: in our case, furnishing a query to
an information system. The use of \emph{domain knowledge} has turned out to
be crucial since our particular goal is to process a queries without any request
of clarification from the system. Due to the inaccuracy and ambiguity generated
by previous phases of analysis we need to select the best hypotheses and often
recover information lost during that selection. There are several ways of integrating
lexical resources (e.g. dictionaries, thesauri) and knowledge bases or ontologies
at different levels of dialogue processing.

\subsection{Robust Definite Clause Grammars}

LHIP (Left-corner Head-driven Island Parser) \cite{ballim94:_lhip,lieske98:rethink_nlp_prolog}
is a system which performs robust analysis of its input, using a grammar defined
in an extended form of the Definite Clause Grammar (DCGs) formalism used for
implementation of parsers in Prolog. LHIP employs a different control strategy
from that used by Prolog DCGs, in order to allow it to cope with ungrammatical
or unforeseen input. A number of tools are provided for producing analyses of
input by the grammar with certain constraints. For example, to find the set
of analyses that provide maximal coverage over the input, to find the subset
of the maximal coverage set that have minimum spans, and to find the analyses
that have maximal thresholds. In addition, other tools can be used to search
the chart for constituents that have been found but are not attached to any
complete analysis.

\paragraph{Weighted LHIP rules}

The main goal of introducing weights into LHIP rules is to induce a partial
order over the generated hypotheses. The following schema illustrates how to
build a simple weighted rule in a compositional fashion where the resulting
weight is computed from the sub-constituents using the minimum operator. Weights
are real numbers in the interval \( [0,1] \). 

\begin{lyxcode}
{\small cat(cat(Hyp),Weight)~\~{}\~{}>~}{\small \par}

{\small ~~~~~~~sub\_cat1(H1,W1),}{\small \par}

{\small ~~~~~~~...,}~\\
{\small ~~~~~~~sub\_catn(Hn,Wn),}~\\
{\small ~~~~~~~\{app\_list({[}H1,...,Hn{]},Hyp),}{\small \par}

{\small ~~~~~~~~min\_list({[}W1,...,Wn{]},Weight)\}.}{\small \par}
\end{lyxcode}
This strategy is not the only possible since the LHIP formalism allows a greater
flexibility. Without entering into formal details we can observe that if we
strictly follow the above schema and we impose a cover threshold of 1 we are
dealing with \emph{fuzzy DCG grammars} \cite{lee69:_note,asveld96:_towar_robus_fuzzif}.
We actually extend this class of grammars with a notion of \emph{fuzzy-robustness}
where weights are used to compute confidence factors for the membership of islands
to categories\footnote{%
Development of this notion is currently under investigation and not yet formalized.
}. Note that this could be useful when we don't want to use deep parsing strategies
and when our goal is to find semantic markers which allow us to segment the
sentence into coarse grain chunks. Furthermore the order of constituents may
play an important role in assigning weights for different rules having the same
number and type of constituents. Each LHIP rule returns a weight together with
a term which will contribute to build the resulting parsing structure. The confidence
factor for a pre-terminal rule is assigned statically on the basis of the domain
knowledge which allows us to find semantic markers within the text.

\subsection{Robust semantic parsing}

In our case study we try to integrate the above principles in order to effectively
compute hypotheses for the query generation task. This can be done by building
a \emph{query hypotheses lattice} and selecting the best ones. The lattice of
hypotheses is generated by means of a LHIP \emph{weighted grammar} extracting
what we called \emph{semantic chunks.} At the end of this process we obtain
suitable interpretations from which we are able to extract the content of the
query. The rules are designed considering two kind of knowledge: \emph{domain
knowledge} is exploited to provide quantitative support (or confidence factor)
to our rules; \emph{linguistic knowledge} is used for determining constraints
in order to prune the hypotheses space. We are concerned with lexical knowledge
when we need to specify lexical LHIP rules which represent the building blocks
of our parsing system. 

\emph{Semantic markers} are domain-dependent word patterns and must be defined
for a given corpus. They identify \emph{cue-phrases} serving both as \emph{separators}
between two logical subparts of the same sentence and as \emph{anchors} for
\emph{semantic constituents.} In our specific case they allow us to search for
the content of the query only in interesting parts of the sentence. The generation
of query hypotheses is performed by: composing weighted rules, assembling semantic
chunks and filtering possible hypotheses.

\paragraph{Lexical knowledge: semantic markers}

As pointed out in \cite{basili97:_lexic_acquis_infor_extrac}, lexical knowledge
plays an important role in Information Extraction since it can contribute in
guiding the analysis process at various linguistic level. In our case we are
concerned with lexical knowledge when we need to specify lexical LHIP rules
which represent the building blocks of our parsing system. \emph{Semantic markers}
are domain-dependent word patterns and must be defined for a given corpus. They
identify \emph{cue-words} serving both as \emph{separators} among logical subparts
of the same sentence and as \emph{introducers} of \emph{semantic constituents.}
In our specific case they allow us to search for the content of the query only
in interesting parts of the sentence. One of the most important separators is
the \emph{announcement-query separator}.

\paragraph{Generation of hypotheses}

The generation of annotation hypotheses is performed by: composing weighted
rules, assembling chunks and filtering possible hypotheses. In this case the
grammar should provide a mean to provide an empty constituent when all possible
hypothesis rules have failed. This is possible using negation and epsilon-rules
in LHIP. The highest level constituent is represented by the whole sentence
structure which simply specifies the possible orders of chunks relative to annotation
hypotheses. In the corresponding rules we have specified a possible order of
chunks interleaved by semantic markers (e.g. separators and introducers). The
computation of global weight may be complex. We simply used the minimum of each
hypothesis confidence values.

\subsection{Filtering and completion}

The obtained frame hypotheses can be further filtered by both using structural
knowledge (e.g. constraints imposed by the syntax analysis) and domain knowledge
(e.g. an ontology like Wordnet). In order to combine the information extracted
from the previous analysis step into the final query representation which can
be directly mapped into the database query language we make use of a frame structure
in which slots represent information units or attributes in the database. We
combine multiple theories representing domain knowledge in order to perform
both consistency checking and the frame completion. A simple notion of context
is used in order to fill by default those slots for which we have no explicit
information. For doing this type of \emph{hierarchical reasoning} we exploit
the meta-programming capabilities of logic programming and we used a meta-interpreter
which allows multiple inheritance among logical theories \cite{BroTur95}.

\section{Conclusions}

So far we have presented a robust speech understanding system that is not far
removed from many other systems. In particular, keyword spotting is a technique
often used in restricted domains. Certainly, we go further by using weighting
techniques on the grammar, employing a logical intermediate representation,
and performing inference on this intermediate representation. The question we
now wish to address, is how can we move forward. Can this approach be generalized?
What are the consequences of this approach? We will argue that this method fits
into a general approach that we call a predictive dialogue modeling approach.
First, however, it is necessary to mix in general remarks about the state of
the art in dialogue processing and the problems that must be addressed. The
advancement from system directed queries to mixed strategies is an important
first stage in allowing for more natural interactive systems. Of course, a mixed
initiative approach typically generates higher error rates. Reducing these error
rates involves constraining dialogues which is typically done by restricting
the domain of application of the system. Such an approach allows us to restrict
the vocabulary to maybe a few hundred words instead of the thousands or hundreds
of thousands of words that we would need in a more general case. An observation
of human to human communication shows a large number of phenomena which present
particular problems for machine analysis. Interruptions, confirmations, anaphora,
ellipsis as well as the breaks, repairs, pauses, and jumps normally found in
human dialogue all present difficulties for machine understanding. Robust processing
goes a long way to handling certain of these problems. We contend, however,
that more general solutions can only come from having a model of the domain
and of the user. The model of the user is not only necessary for better understanding
what the user is saying, but also for matching the expectations of the user
in the interaction with the machine. This is necessary because it is difficult
to communicate the system's capabilities to the user. The user does not necessarily
know the vocabulary that the system's capable of handling, nor the type of questions
that the system may answer. We can see then that a user model can be of great
benefit in future natural interactive systems. In addition, in multi-modal interaction
the user model will allow us to better tailor the use of different modalities
to the user. More importantly, from our point of view, such a model is part
of a predictive approach to natural interactivity. 

The idea of this approach is to continuously anticipate the interaction with
the user. In other words, analysis should be based on the expectations of the
system. Such an approach allows us to restrict vocabulary, domain knowledge,
and interaction types to only those necessary for the immediate understanding.
In a sense dialogue grammars, finite state approaches to dialogue, and template
approaches to dialogue are all predictive models. We anticipate an approach
in which more general models of language, based on the content of communication,
are derived from knowledge of the domain, the user's knowledge of the domain,
and the system's view of the user's needs, beliefs, goals and motivations.

\subsection{Related works }

As examples of robust approaches applied to dialogue systems we cite here two
systems which are based on similar principles. In the DIALOGOS human-machine
telephone system (see \cite{albesano97:_dialog}) the robust behavior of the
dialogue management module is based both on a contextual knowledge base of pragmatic-based
expectations and the dialogue history. The system identifies discrepancies between
expectations and the actual user behavior and in that case it tries to rebuild
the dialogue consistency. Since both the domain of discourse and the user's
goals (e.g. railway timetable inquiry) are clear, it is assumed the systems
and the users cooperate in achieving reciprocal understanding. Under this underlying
assumption the system pro-actively asks for the query parameters and it is able
to account for those spontaneously proposed by the user. 

In the SYSLID project \cite{boros96:_proces_spoken_dialog_system_exper_syslid_projec}
where a robust parser constitutes the linguistic component of the query-answering
dialogue system. An utterance is analyzed while at the same time its semantical
representation is constructed. This semantical representation is further analyzed
by the dialogue control module which then builds the database query. Starting
from a word graph generated by the speech recognizer module, the robust parser
will produce a search path into the word graph. If no complete path can be found,
the robust component of the parser, which is an island based chart parser \cite{hanrieder95:_robus_parsin_spoken_dialog_using},
will select the maximal consistent partial results. In this case the parsing
process is also guided by a lexical semantic knowledge base component that helps
the parse in solving structural ambiguities.

\subsection{Future Work }

The limited resources of the project did not allow us to adequately evaluate
the results and test the system against real situations. Nonetheless our final
opinion about the ISIS project is that there are some promising directions applying
robust parsing techniques and integrating them with knowledge representation
and reasoning. Moreover we did not commit on the used architecture and we envision
that better results can be achieved moving towards a distributed agent-based
architecture for natural language processing. An ongoing project\footnote{%
More information about the HERALD (Hybrid Environment for Robust Analysis of
Language Data) project are available at: \texttt{http://lithwww.epfl.ch/\~{}pallotta/herald.ps.gz}
} at our laboratory is concerned with these aspects, where we propose an hybrid
distributed architecture which combines symbolic and numerical computing by
means of agents providing linguistic services. Within this architecture also
the knowledge management plays a central role and it is aimed to the intelligent
coordination of the linguistic agents \cite{ballim91:_artif_believ, ballim93:_propos_attit_framew_requir}. 

{\footnotesize \bibliographystyle{plain}

}{\footnotesize \par}

\end{document}